\pdfoutput=1

\documentclass[11pt]{article}

\usepackage[]{EMNLP2023}

\usepackage{times}
\usepackage{latexsym}

\usepackage[T1]{fontenc}

\usepackage[utf8]{inputenc}

\usepackage{microtype}

\usepackage{inconsolata}

\usepackage{amsmath}
\usepackage{times}
\usepackage{framed}
\usepackage{ragged2e}
\usepackage{bm}
\usepackage{soul}
\usepackage{latexsym}
\usepackage{subfigure}
\usepackage{amsthm}
\usepackage{graphicx}
\usepackage{float}
\usepackage{longtable}
\usepackage{booktabs} 
\usepackage{multirow}
\usepackage{multicol}
\usepackage{amssymb}
\usepackage{dashbox}%
\usepackage{multirow}
\usepackage{textcomp}
\usepackage{tcolorbox}
\usepackage{makecell}
\usepackage{bbding}
\usepackage[ruled, vlined, linesnumbered]{algorithm2e}
\usepackage{mathtools}
\usepackage{adjustbox}
\usepackage[ruled,vlined]{algorithm2e}
\usepackage{color, soul}
\usepackage{pifont}
\usepackage{listings}
\usepackage{array}
\usepackage{tabularx}
\newcolumntype{L}{>{\RaggedRight\hangafter=1\hangindent=0em}X}
\newcolumntype{P}[1]{>{\centering\arraybackslash}p{#1}}
\newcolumntype{M}[1]{>{\centering\arraybackslash}m{#1}}
\usepackage{cleveref}
\crefname{section}{§}{§§}
\Crefname{section}{§}{§§}
\crefname{figure}{Figure}{Figure}
\Crefname{figure}{Figure}{Figure}
\crefname{table}{Table}{Table}
\Crefname{table}{Table}{Table}
\definecolor{my_red}{RGB}{255,99,71}
\definecolor{my_green}{RGB}{50,205,50}
\definecolor{my_blue}{RGB}{65,105,225}
\definecolor{forestgreen}{HTML}{228B22}

\newcommand\Rbase{RoBERTa$_{\textsc{BASE}}$\xspace}
\newcommand\ourtool{OmniEvent\xspace}
\newcommand\mycheck{\textcolor{forestgreen}{\Checkmark}\xspace}
\newcommand\myx{\textcolor{red}{\XSolidBrush}\xspace}

\definecolor{codegreen}{rgb}{0,0.6,0}
\definecolor{codegray}{rgb}{0.5,0.5,0.5}
\definecolor{codepurple}{rgb}{0.58,0,0.82}
\definecolor{backcolour}{rgb}{0.95,0.95,0.92}

\lstdefinestyle{mystyle}{
    backgroundcolor=\color{backcolour},   
    commentstyle=\color{codegreen},
    stringstyle=\color{codepurple},
    basicstyle=\ttfamily\scriptsize,
    breakatwhitespace=true,         
    breaklines=true,                 
    captionpos=b,                    
    keepspaces=true,                 
    numbers=none,                    
    numbersep=5pt,                  
    showspaces=false,                
    showstringspaces=false,
    showtabs=false,                  
    tabsize=2,
    columns=flexible,
    escapeinside={(*}{*)},
}

\title{\includegraphics[width=3.0cm]{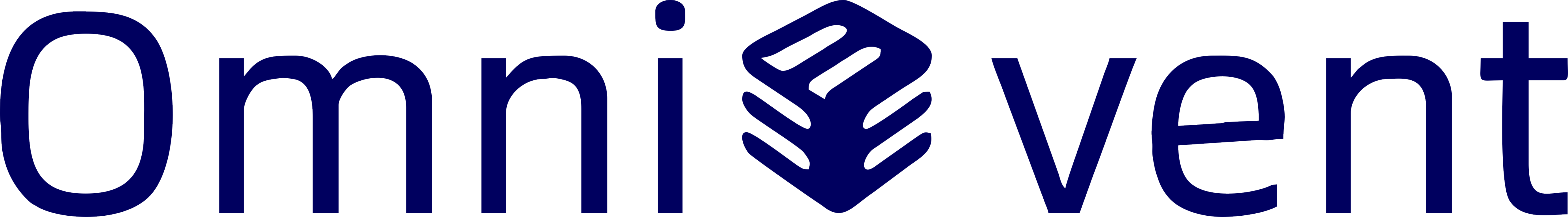}: A Comprehensive, Fair, and Easy-to-Use Toolkit for \\ Event Understanding}

\author{Hao Peng\thanks{\quad Equal contribution.}\hspace{0.5em}, Xiaozhi Wang$^{*}$, 
Feng Yao, Zimu Wang, \\ \textbf{Chuzhao Zhu, Kaisheng Zeng, Lei Hou, Juanzi Li} \\
Tsinghua University 
}

\begin{document}
\maketitle
\begin{abstract}
Event understanding aims at understanding the content and relationship of events within texts, which covers multiple complicated information extraction tasks: event detection, event argument extraction, and event relation extraction. To facilitate related research and application, we present an event understanding toolkit \ourtool, which features three desiderata: (1) \textbf{Comprehensive.} \ourtool supports mainstream modeling paradigms of all the event understanding tasks and the processing of $15$ widely-used English and Chinese datasets. (2) \textbf{Fair.} \ourtool carefully handles the inconspicuous evaluation pitfalls reported in \citet{peng2023devil}, which ensures fair comparisons between different models. (3) \textbf{Easy-to-use}. \ourtool is designed to be easily used by users with varying needs. We provide off-the-shelf models that can be directly deployed as web services. The modular framework also enables users to easily implement and evaluate new event understanding models with \ourtool. The toolkit\footnote{\url{https://github.com/THU-KEG/OmniEvent}} is publicly released along with the demonstration website and video\footnote{\url{https://omnievent.xlore.cn/}}.

\end{abstract}

\section{Introduction}
\begin{figure}[!t]
    \centering
    \includegraphics[width=\linewidth]{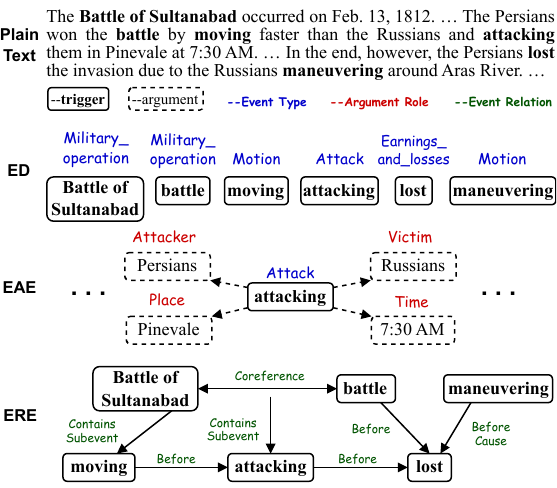}
    \caption{An illustration for the event understanding tasks, including event detection (ED), event argument extraction (EAE), and event relation extraction (ERE).}
    \label{fig:pipeline}
\end{figure}




Correctly understanding events is fundamental for humans to understand the world. Event understanding requires identifying real-world events mentioned in texts and analyzing their relationships, which naturally benefits various downstream applications, such as stock prediction~\citep{ding2015deep}, adverse drug event detection~\citep{wunnava2019adverse}, narrative event prediction~\citep{wang2021incorporating}, and legal case analysis~\citep{leven}. 

As illustrated in Figure~\ref{fig:pipeline}, event understanding covers three complicated information extraction tasks: (1) event detection (ED), which is to detect the event triggers (keywords
or phrases evoking events in texts) and classify their event types, (2) event argument extraction (EAE), which is to extract the event arguments for each trigger and classify their argument roles, and (3) event relation extraction (ERE), which is to identify the complex relationships between events, typically including temporal, causal, coreference, and subevent relations. ED and EAE together constitute the conventional event extraction (EE) task.

In recent years, event understanding research has grown rapidly~\citep{ma2022prompt, wang2022maven, yue2023zero, huang2023more}, and multiple practical systems~\citep{wadden2019entity,lin2020joint,zhang2020openue,du2022resin,zhang2022deepke} have been developed.
However, as shown in Table~\ref{tab:comparison}, existing systems exhibit several non-negligible issues: (1) \textbf{Incomprehensive Tasks}. Existing systems mainly focus on the two EE subtasks and rarely cover the whole event understanding pipeline with ERE tasks. The notable exception EventPlus~\citep{eventplus} merely covers the temporal relations. (2) \textbf{Limited Support for Redevelopment and Evaluation}. Most of the existing event understanding systems are highly integrated and not extensible, which means users cannot easily develop new models within their frameworks. Especially considering the recent rise of large language models (LLMs)\footnote{The definition of LLM is vague. Here we use ``LLM'' to refer to models with more than 10 billion parameters.}, adequate support for LLMs is urgent but often missing. Moreover, the complicated data processing and evaluation details often lead to inconsistent and unfair evaluation results~\citep{peng2023devil}, but existing systems do not pay much attention to evaluations.

To address these issues, we develop \ourtool, a comprehensive, fair, and easy-to-use toolkit for event understanding, which has three main features:
(1) \textbf{Comprehensive Support for Task, Model, and Dataset.} \ourtool supports end-to-end event understanding from plain texts, i.e., all the ED, EAE, and ERE tasks. For ED and EAE, we classify the mainstream methods into four paradigms, including classification, sequence labeling, span prediction, and conditional generation. We implement various representative methods for each paradigm. For ERE, we provide a unified modeling framework and implement a basic pairwise classification method~\citep{wang2022maven}. We also cover the preprocessing of $15$ widely-used English and Chinese datasets.  
(2) \textbf{Fair Evaluation.} As found in~\citet{peng2023devil}, there are three major pitfalls hidden in EE evaluation, including data processing discrepancy, output space discrepancy, and absence of pipeline evaluation. \ourtool implements all the proposed remedies to help users avoid them. Specifically, we implement unified pre-processing for all the datasets and a method to convert the predictions of different paradigms into a unified space. \ourtool also provides unified prediction triggers of supported datasets for fair pipeline comparisons. 
(3) \textbf{Easy-to-Use for Various Needs.} We design a modular and extensible framework for \ourtool, which appeals to users with various needs. We provide several off-the-shelf models that can be easily deployed and used by users interested in applications. Model developers and researchers can train implemented methods within several lines of code or customize their own models and evaluate them. By integrating Transformers~\citep{wolf-etal-2020-transformers} and DeepSpeed~\citep{deepspeed}, \ourtool also supports efficiently fine-tuning LLMs as backbones.

To demonstrate the effectiveness of OmniEvent, we present the results of several implemented methods on widely-used benchmarks. We also conduct experiments with models at different scales and show that fine-tuning LLMs helps achieve better event understanding results. We hope \ourtool could facilitate the research and applications of event understanding.

\section{Related Work}
\begin{table}[t]
    \centering
    \small
    \begin{adjustbox}{max width=1\linewidth}
{
    \begin{tabular}{l|ccccc}
    \toprule
    System & EE & ERE & \makecell[c]{\#Supported \\ Models} & \makecell[c]{\#Supported \\ Datasets} & \makecell[c]{LLM \\ Support} \\
    \midrule
    DYGIE & \mycheck  & \myx & $1$  & $1$  & \myx \\
    OneIE & \mycheck  & \myx & $1$ & $4$  &\myx \\
    OpenUE & \mycheck  & \myx & $1$ & $2$  &\myx \\
    EventPlus & \mycheck  & \mycheck & $1$ & N/A  &\myx \\
    FourIE  & \mycheck  & \myx & $1$ & N/A  &\myx \\
    RESIN-11 & \mycheck  & \myx & $1$ & N/A  &\myx \\
    DeepKE & \mycheck  & \myx & $2$ & $1$  &\mycheck \\
    \midrule
    OmniEvent & \mycheck  & \mycheck & >$20$ & $15$  &\mycheck \\
    \bottomrule
    \end{tabular}
}
\end{adjustbox}
    \caption{Comparisons between OmniEvent and other event understanding systems. The number of supported models and datasets only includes those of event understanding tasks. N/A denotes that the system is an integrated service and does not process benchmark datasets. For OmniEvent, the module combination enables many possible models and $20$ is the number of models we have tested for usability.}
    \label{tab:comparison}
\end{table}

\begin{figure*}[ht]
    \centering
    \includegraphics[width=0.99\linewidth]{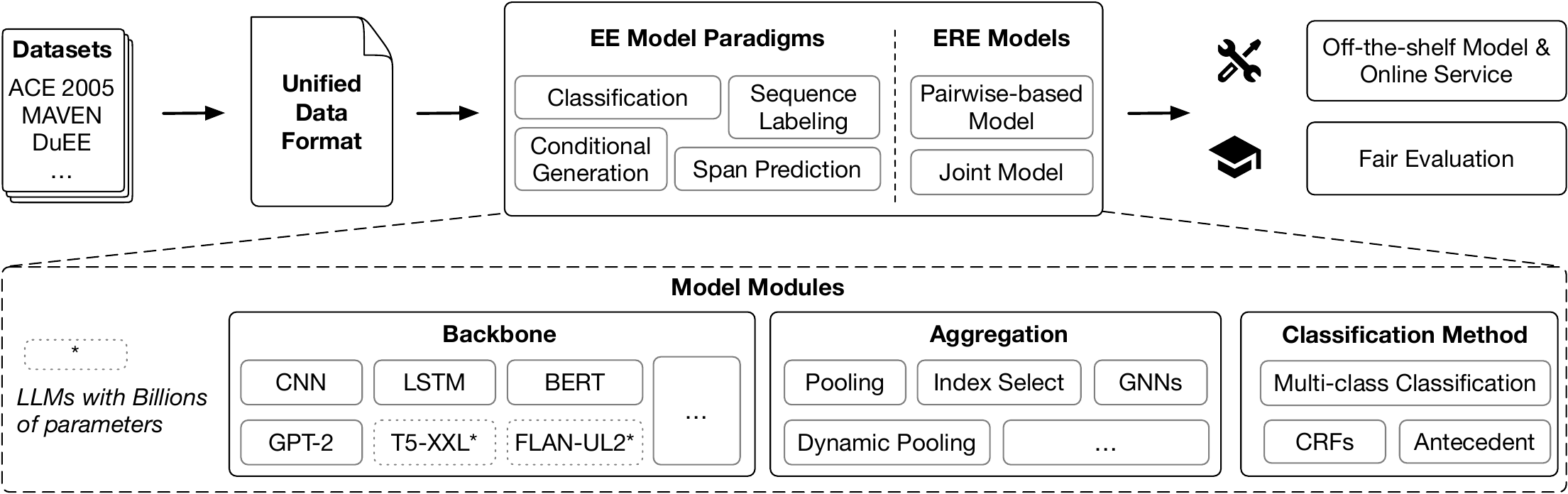}
    \caption{Overview of the OmniEvent toolkit. OmniEvent can serve as a system offering event understanding services to users, while also serving as a toolkit for researchers in model development and evaluation.
    OmniEvent provides pre-processing scripts for widely-used datasets and converts the datasets into a unified data format. OmniEvent provides modular components and users can easily develop a new model based on the components. OmniEvent also supports large language models (T5-XXL~\citep{t5} and FLAN-UL2~\citep{ul2}).}
    \label{fig:overview}
\end{figure*}

With the advancement of research in NLP, various toolkits or systems for event understanding have been developed. They tend to focus on developing advanced EE systems to achieve improved results on public benchmarks~\citep{wadden2019entity, lin2020joint, FourIE} or  
perform robustly in real-world scenarios~\citep{newsreader, du2022resin}. However, these toolkits or systems, designed based on a specific EE model, do not support comprehensive implementations of EE models and are inconvenient for secondary development. There is also some work that has meticulously designed user-friendly algorithmic frameworks~\citep{zhang2020openue, zhang2022deepke}, which are convenient for usage and secondary development. However, they are not specifically designed for event understanding, hence the corresponding support is limited. EventPlus~\citep{eventplus} is the only work supporting the entire event understanding pipeline but it only supports temporal relation extraction and does not provide comprehensive implementations of event understanding models. Moreover, existing work also neglects the discrepancies in EE evaluation as mentioned in~\citet{peng2023devil}, which may result in unfair comparison. Finally, in the era of LLMs, existing work (except for DeepKE) also lacks support for LLMs.

Considering the mentioned issues, we present OmniEvent, a comprehensive, fair, and easy-to-use toolkit for event understanding. Compared to other systems in Table~\ref{tab:comparison}, OmniEvent supports the entire event understanding pipeline and comprehensively implements various models. OmniEvent also supports efficient fine-tuning and inference of LLMs. Meanwhile, OmniEvent provides respective remedies for eliminating the discrepancies as mentioned in~\citet{peng2023devil}. With a modular implementation and several released off-the-shelf models, OmniEvent is user-friendly and easy to use.

\section{The OmniEvent Toolkit}
We introduce the overview (\cref{sec:overview}) and main features of \ourtool (\cref{sec:comprehensive_implementation,sec:fair_evaluation,sec:easy_to_use}), as well as an online demonstration (\cref{sec:online_system}) powered by \ourtool.

\subsection{Overview}
\label{sec:overview}

The overall architecture of OmniEvent is illustrated in Figure~\ref{fig:overview}. OmniEvent provides a data pre-processing module for unified pre-precessing. Users can either use the supported datasets or customize their own datasets. 
After pre-processing, OmniEvent provides a flexible modular framework for model implementation. OmniEvent abstracts and disassembles the mainstream models into three basic modules and implements the basic modules in a highly encapsulated way. By combining our provided modules or implementing their own modules, users can easily assemble a model. OmniEvent reproduces several widely-used models in this way. Finally, OmniEvent provides a fair evaluation protocol to convert predictions of different models into a unified and comparable output space. 

\subsection{Comprehensive Support}
\label{sec:comprehensive_implementation}
OmniEvent implements the entire event understanding pipeline, i.e., all the ED, EAE, and ERE tasks, and can serve as a one-stop event understanding platform. Furthermore, OmniEvent provides
comprehensive coverage of models and datasets.
\paragraph{Models}
OmniEvent comprehensively implements representative models for ED, EAE, and ERE. For ED and EAE, OmniEvent covers four mainstream method paradigms, which contain: (1) classification methods, including DMCNN~\citep{chen2015event}, DMBERT~\citep{wang-etal-2019-adversarial-training}, and CLEVE~\citep{wang-etal-2021-cleve}, which classify event or argument candidates into appropriate types, (2) sequence labeling methods, including BiLSTM+CRF~\citep{wang-etal-2020-maven} and BERT+CRF~\citep{wang-etal-2020-maven}, which labels the sequences with the BIO format, (3) span prediction method, including EEQA~\citep{du-cardie-2020-event}, which predicts the boundaries of event and argument spans, (4) conditional generation method, including Text2Event~\citep{lu-etal-2021-text2event}, which directly generates the answers. Moreover, as shown in Figure~\ref{fig:overview}, OmniEvent implements various basic modules and the users can easily combine different modules to build new models, e.g., combining GPT-2~\citep{radford2019language} and CRF~\citep{crf} (GPT-2+CRF). For event relation extraction, OmniEvent implements a unified pairwise relation extraction framework. Especially for the event coreference resolution task, OmniEvent develops an antecedent ranking method. As extracting different relations (causal, temporal) may benefit each other~\citep{wang2022maven}, we develop a joint event relation extraction model in OmniEvent. 

\paragraph{Datasets}
\begingroup
\begin{table}
    \centering
    \small
    \begin{tabular}{lp{0.8\linewidth}}
    \toprule
    EE & ACE 2005~\citep{walker2006ace}, TAC KBP~\citep{ellis2014overview, ellis2015overview, ellis2016overview, getman2017overview}, RichERE~\citep{song2015light}, MAVEN~\citep{wang-etal-2020-maven}, \textit{ACE 2005 (zh)}~\citep{walker2006ace}, \textit{LEVEN}~\citep{leven}, \textit{DuEE}~\citep{li2020duee}, \textit{FewFC}~\citep{Yang2021fewfc} \\
    \midrule
    ERE & MAVEN-ERE~\citep{wang2022maven}, ACE 2005~\citep{walker2006ace}, TB-Dense~\citep{chambers-etal-2014-dense}, MATRES~\citep{ning2018matres}, TCR~\citep{ning-etal-2018-joint}, CausalTB~\citep{mirza-etal-2014-annotating}, EventStoryLine~\citep{caselli-vossen-2017-event}, HiEve~\citep{glavas-etal-2014-hieve} \\
    \bottomrule
    \end{tabular}
    \caption{Currently supported datasets in OmniEvent. \textit{Italics} represent Chinese datasets.}
    \label{tab:datasets}
\end{table}
\endgroup


As shown in Table~\ref{tab:datasets}, OmniEvent includes various widely-used Chinese and English event understanding datasets, covering general, legal, and financial domains. For each included dataset, we provide a pre-processing script to convert the dataset into a unified format, as shown in appendix~\ref{sec:appendix_unfieid_data_format}. For datasets with different pre-processing scripts, e.g., ACE 2005, OmniEvent provides all the mainstream scripts for users.

\subsection{Fair Evaluation}
\label{sec:fair_evaluation}
\begin{figure}
\begin{minipage}{\linewidth}
\begin{lstlisting}[language=Python, label=code, caption={Example for converting the sequence labeling, span prediction, and conditional generation predictions into a unified output space.}]
from OmniEvent import convert_SL, convert_SP, convert_CG

text = "City A suffers a terrorist attack in 2021 ."
tokens = text.split()
events = [{
    "type": "attack",
    "trigger": "terrorist attack",
    "offset": [4, 6]
}]

# predictions generated by users
predictions_SL = [O, O, O, B-Attack, I-Attack, I-Attack, O, O, O]
# obtain comparable results
results = convert_SL([predictions_SL], [events], [tokens])

predictions_SP = [{"offset": [3, 6], "type": "attack"}]
results = convert_SP([predictions_SP], [events], [tokens])

# without offsets
predictions_CG = [{"trigger": "a terrorist attack", "type": "attack"}]
results = convert_CG([predictions_CG], [events], [tokens])
\end{lstlisting}
\end{minipage}
\end{figure}

As discussed in~\citet{peng2023devil}, there exist several pitfalls in EE evaluation that significantly influence the fair comparison of different models. They are in three aspects: data-preprocessing discrepancy, output space discrepancy, and absence of pipeline evaluation. OmniEvent proposes remedies for eliminating them. 
\paragraph{Specify data pre-processing} As the data pre-processing discrepancy mainly comes from using different processing options, OmniEvent provides all the widely-used data pre-processing scripts. Users only need to specify the pre-processing script for comparable results with previous studies. 
\paragraph{Standardize output space} As suggested in~\citet{peng2023devil}, OmniEvent provides several easy-to-use functions to convert the predictions of different models into a unified output space. Code~\ref{code} shows the conversion codes of sequence labeling, span prediction, and conditional generation predictions for event detection. Users can easily utilize the functions to obtain fair and comparable results. 

\paragraph{Pipeline evaluation} The pipeline evaluation requires conducting EAE based on predicted triggers. Therefore, the results of EAE models are comparable only when using the same predicted triggers. OmniEvent provides a unified set of predicted triggers for widely-used datasets. Specifically, OmniEvent leverages CLEVE~\citep{wang-etal-2021-cleve}, an advanced ED model, to predict triggers for widely-used EE datasets: ACE 2005, KBP 2016, KBP 2017, and RichERE.

\subsection{Easy-to-Use}
\label{sec:easy_to_use}

\begin{figure}
\begin{minipage}{\linewidth}
\begin{lstlisting}[language=Python, label=code2, caption={Example of using inference interface and off-the-shelf models for event understanding.}]
from OmniEvent.infer import infer
# input text
text = "U.S. and British troops were moving on the strategic southern port city of Basra Saturday after a massive aerial assault pounded Baghdad at dawn"
# event detection
ed_results = infer(text=text, task="ED")
# end-to-end event extraction
ee_results = infer(text=text, task="EE")
# end-to-end event understanding 
# event extraction & relation extraction
all_results = infer(text=text, task="EE & ERE")
\end{lstlisting}
\end{minipage}
\end{figure}

OmniEvent is designed to be user-friendly and easy to use. 
Specifically, OmniEvent incorporates the following designs.
\paragraph{Easy start with off-the-shelf models}
OmniEvent provides several off-the-shelf models for event understanding. Specifically, we train a multilingual T5~\citep{mt5} for ED and EAE on the collection of included EE datasets, respectively. And we train a joint ERE model based on RoBERTa~\citep{roberta} on the training set of MAVEN-ERE. As shown in Code~\ref{code2}, OmniEvent provides an interface for inference and users can easily use these models in their applications with a few lines of code.
\paragraph{Modular implementation}
As shown in Figure~\ref{fig:overview}, OmniEvent abstracts and disassembles the mainstream models into basic modules. The backbone module implements various text encoders, such as CNN~\citep{cnn} and BERT~\citep{bert}, to encode plain texts into low-dimension dense vectors. The backbone module also supports LLMs such as T5-XXL~\citep{t5} and FLAN-UL2~\citep{ul2}. The aggregation module includes various aggregation operations, which aggregate and convert the dense vectors into representations of events, arguments, and relations. The classification module projects the representations into distributions of classification candidates. 
With the highly modular implementation, users can easily combine the basic modular components to develop new models.
\paragraph{Efficient support for LLMs} OmniEvent is built upon Huggingface's Transformers~\citep{wolf-etal-2020-transformers} and DeepSpeed~\citep{deepspeed}, an efficient deep learning optimization library. With the built-in DeepSpeed support, OmniEvent can be used to train and infer LLMs efficiently with only modifications of the startup shell scripts.



\subsection{Online Demonstration}
\label{sec:online_system}

\begin{figure}
    \centering
    \includegraphics[width=\linewidth]{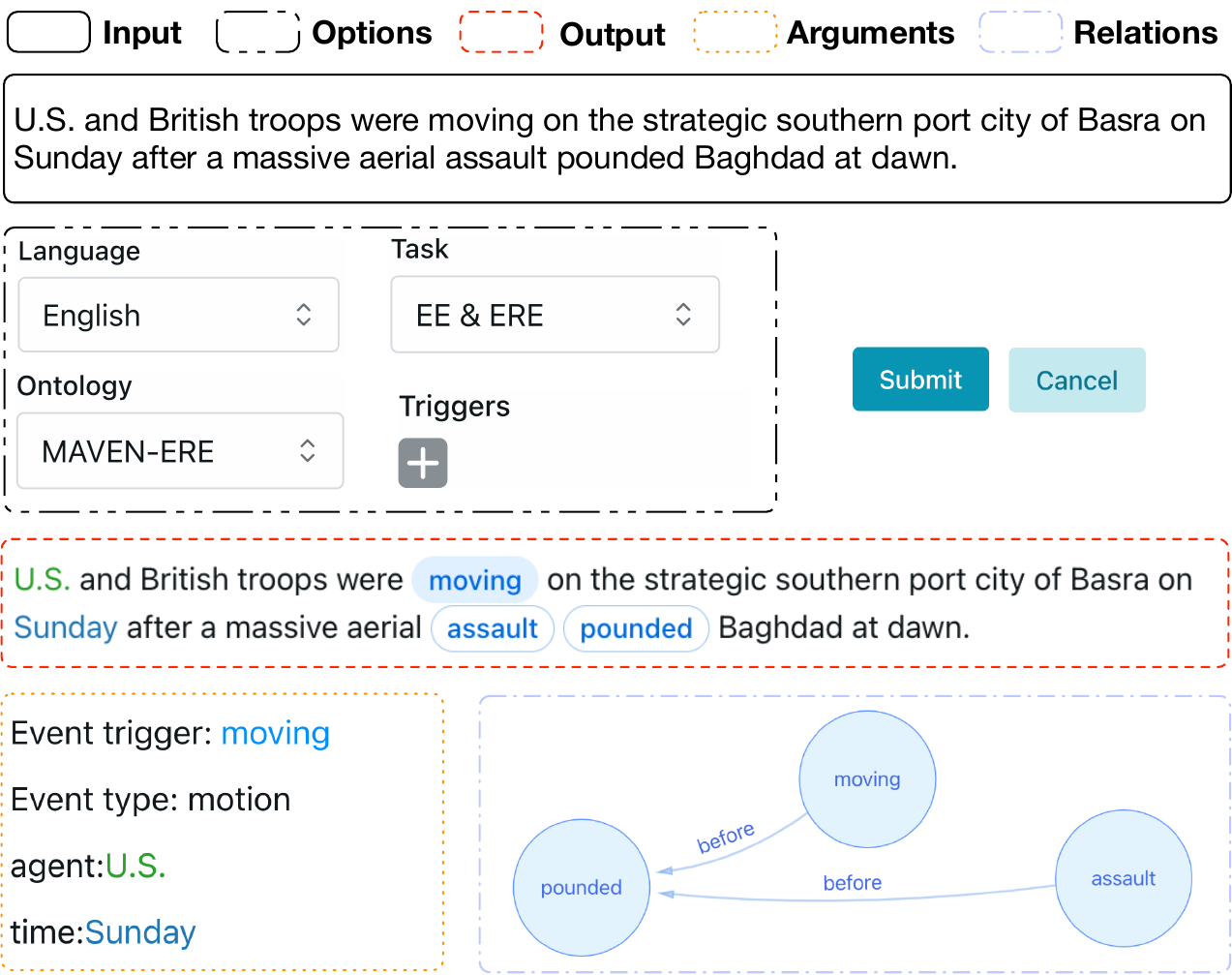}
    \caption{Example of the online demonstration. We re-arrange the layout of the website for a compact presentation. Better visualization in color.}
    \label{fig:website}
\end{figure}

Besides the OmniEvent toolkit, we also develop an online demonstration system\footnote{\url{https://omnievent.xlore.cn/}} powered by OmniEvent. We train and deploy a multilingual T5$_{\textsc{BASE}}$ model for EE and a RoBERTa$_{\textsc{BASE}}$ model for event relation extraction. The website example is shown in Figure~\ref{fig:website}. The online system supports EE based on various English and Chinese classification schemata and ERE based on the MAVEN-ERE schema. The website mainly contains three parts. The input part includes a text entry field and several options. Users can choose the language, task, and ontology (i.e., classification schema) for event understanding. The results of EE are shown in the output field with extracted triggers and arguments highlighted. The results of ERE are shown as an event knowledge graph, where a node is an event and an edge is an identified relation between events. The example in Figure~\ref{fig:website} shows the results of end-to-end event understanding (ED, EAE, and ERE) from the input plain text.
\section{Evaluation}
In this section, we conduct empirical experiments to evaluate the effectiveness of the OmniEvent toolkit on widely-used datasets.

\subsection{Event Extraction}

    

\begin{table}
    \centering
    \small
    \begin{adjustbox}{max width=1\linewidth}
{
    \begin{tabular}{llcccc}
    \toprule
    Task & Dataset & CLS & SL & SP & CG \\
    \midrule
    \multirow{6}{*}{ED} & ACE 2005 & $68.6$ & $68.6$ & $71.0$ & $66.0$ \\
    & RichERE & $51.4$ & $50.1$ & $50.4$ & $51.4$ \\
    & MAVEN & $68.6$ & $68.6$ & $68.1$ & $61.9$ \\
    & ACE 2005 (ZH) & $75.8$ & $75.9$ & $73.5$ & $71.6$ \\
    & LEVEN & $85.2$ & $84.7$ & $84.3$ & $81.4$ \\ 
    & FewFC & $67.2$ & $62.3$ & $59.0$ & $71.3$ \\
    \midrule
    \multirow{4}{*}{EAE} & ACE 2005 & $58.7$ & $49.4$ & $40.1$ & $45.7$ \\
    & RichERE & $68.3$ & $59.7$ & $24.3$ & $24.9$ \\
    & ACE 2005 (ZH) & $73.1$ & $67.9$ & $35.4$ & $49.0$ \\
    & FewFC & $68.7$ & $59.8$ & $46.7$ & $53.7$ \\
    \bottomrule
    \end{tabular}
}
\end{adjustbox}
    \caption{Experimental results (F1,\%) of implemented EE models in OmniEvent on various EE datasets. CLS: Classification; SL: Sequence labeling; SP: Span prediction; CG: Conditional generation. We evaluate the representative models: DMBERT, BERT+CRF, EEQA, and Text2Event for CLS, SL, SP, and CG, respectively.}
    \label{tab:ee_results}
\end{table}

We evaluate the performance of representative EE models implemented in OmniEvent on various widely-used datasets. All the models are evaluated using the unified evaluation protocol, i.e., the output space is standardized and the results of EAE are from pipeline evaluation. The pre-processing script for ACE 2005 is the same as in~\citet{wadden2019entity}. For EEQA, we utilize the same prompts as in the original paper for ACE 2005 and manually curate prompts for all the other datasets. The results of event detection and event argument extraction are shown in Table~\ref{tab:ee_results}. The results demonstrate the effectiveness of OmniEvent, which achieves similar performance compared to their original implementations. OmniEvent provides all the experimental configuration files in the YAML format, which records all the hyper-parameters. Users can easily reproduce the results using the corresponding configuration files.

\subsection{Event Relation Extraction}


\begin{table}[t]
    \centering
    \small
    \begin{adjustbox}{max width=1\linewidth}
{
    \begin{tabular}{llccc}
    \toprule
    Relation Type & Dataset & P & R & F1 \\
    \midrule
    \multirow{2}{*}{Coreference} & ACE 2005 & $94.5$ & $81.7$ & $87.7$ \\
    & MAVEN-ERE & $97.9$ & $98.5$ & $98.2$ \\
    \midrule
    \multirow{4}{*}{Temporal} & TB-Dense & $67.9$ & $54.0$ & $60.2$ \\
    & MATRES & $87.2$ & $93.8$ & $90.4$ \\
    & TCR & $78.3$ & $78.3$ & $78.3$ \\
    & MAVEN-ERE & $53.3$ & $61.4$ & $57.1$ \\
    \midrule
    \multirow{3}{*}{Causal} & CausalTB & $100.0$ & $50.0$ & $66.7$ \\
    & EventStoryLine & $19.5$ & $25.8$ & $22.2$ \\
    & MAVEN-ERE & $36.0$ & $26.4$ & $30.5$ \\
    \midrule
    \multirow{2}{*}{Subevent} & HiEve & $21.4$ & $13.4$ & $16.5$ \\
    & MAVEN-ERE & $30.8$ & $24.3$ & $27.1$ \\
    \bottomrule
    \end{tabular}
}
\end{adjustbox}
    \caption{Experimental results (\%) of the implemented pairwise-based ERE model in OmniEvent on various ERE datasets. The backbone is \Rbase. The evaluation metric for coreference is B-cubed~\citep{b-cubed}.}
    \label{tab:ere_results}
\end{table}

We also conduct empirical experiments to evaluate the performance of ERE models developed in OmniEvent on various widely-used datasets. As shown in Table~\ref{tab:ere_results}, the results are on par or slightly better than the originally reported results in~\citet{wang2022maven}, which demonstrates the validity of ERE models in OmniEvent. We also provide configuration files containing all the hyper-parameter settings for reproduction.

\subsection{Experiments using LLMs}
\begin{figure}
    \centering
    \includegraphics[width=0.9\linewidth]{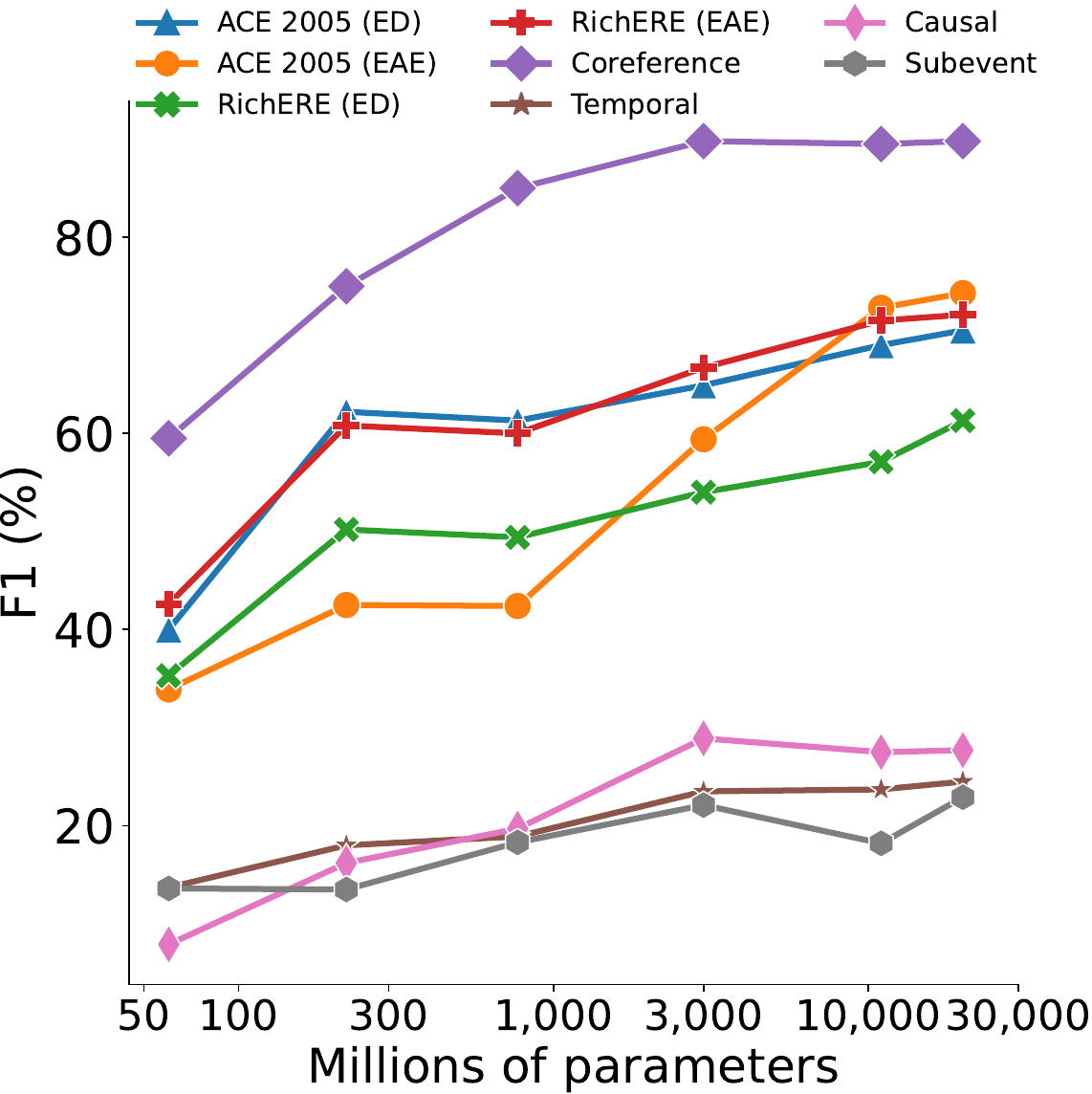}
    \caption{Experimental results of models at different scales on all event understanding tasks.}
    \label{fig:scaling_law}
\end{figure}

OmniEvent supports efficient fine-tuning and inference for LLMs. To examine the effectiveness and validity of LLMs support in OmniEvent and investigate the performance of models at different scales, we train a series of models on several datasets. Specifically, for ED and EAE, we fine-tune FLAN-T5~\citep{flan-t5} (from Small to XXL) and FLAN-UL2~\cite{ul2}, an LLM with 20 billion parameters on ACE 2005 and RichERE. For ERE, due to the lack of encoder-only LLMs, we use the same models as ED and EAE. We convert the ERE task into a sequence generation task. All the experiments are run on Nvidia A100 GPUs. Fine-tuning FLAN-UL2 on ACE 2005 consumes only about 25 GPU hours, which demonstrates the efficiency of LLMs support in OmniEvent. 
The results are shown in Figure~\ref{fig:scaling_law}. We can observe that larger models perform better and FLAN-UL2 achieves remarkable performance on ACE 2005 and RichERE datasets, which demonstrates the validity of LLMs support in OmniEvent. We can also notice that the results of ERE are much worse than the results in Table~\ref{tab:ere_results}, which may be due to the extremely long contexts and complex output space of the ERE task. We hope the findings based on OmniEvent can inspire future research on how to better leverage LLMs for event understanding.
\section{Conclusion and Future Work}
In the paper, we present OmniEvent, a comprehensive, fair, and easy-to-use toolkit for event understanding. With the comprehensive and modular implementation, OmniEvent can help researchers and developers conveniently develop and deploy models. OmniEvent also releases several off-the-shelf models and deploys an online system for enhancing the applications of event understanding models. In the future, we will continually maintain OmniEvent to support more models and datasets and release more effective models.

\section*{Limitations}
The major limitations of OmniEvent are two-fold: (1) OmniEvent currently does not support document-level event extraction models and datasets, such as RAMS~\citep{rams} and WikiEvents~\citep{wikievents}. OmniEvent also lacks support for a wider range of ERE models, such as constrained loss~\citep{WangCZR20joint} and ILP inference~\citep{HanNP19Joint}.  In the future, we will continue to maintain OmniEvent to support a broader range of models and datasets. (2) OmniEvent currently only supports two languages, Chinese and English, and does not yet support event relation extraction in Chinese. This might constrain the widespread usage of the OmniEvent toolkit. In the future, OmniEvent will support more languages. 

\section*{Ethical Considerations}
We will discuss the ethical considerations and broader
impact of this work here: (1) \textbf{Intellectual property.} OmniEvent is open-sourced and released under MIT license\footnote{\url{https://opensource.org/license/mit}}. We adhere to the original licenses for all datasets and models used. Regarding the issue of data copyright, we do not provide the original data and we only provide processing scripts for the original data. (2) \textbf{Environmental Impact.}  The experiments are conducted on the Nvidia A100 GPUs and consume approximately 350 GPU hours. This results in a substantial amount of carbon emissions, which incurs a negative influence on our environment~\citep{strubell-etal-2019-energy}. (3) \textbf{Intended Use.} OmniEvent can be utilized to provide event understanding services for users, and it can also serve as a toolkit to assist researchers in developing and evaluating models. (4) \textbf{Misuse risks.} OmniEvent \textbf{should not} be utilized for processing and analyzing sensitive or uncopyrighted data. The output of OmniEvent is determined by the input text and \textbf{should not} be used to support financial or political claims.

\bibliography{anthology,custom}
\bibliographystyle{acl_natbib}

\appendix
\clearpage
\section*{Appendices}
\section{Unified Data Format}
\label{sec:appendix_unfieid_data_format}

An instance converted to the unified data format is shown in Code~\ref{code_format}. The data format comprehensively records all the event-related information: triggers, arguments, and coreference, temporal, causal, and subevent relations.

\begin{figure}
\begin{minipage}{1.0\textwidth}
\begin{lstlisting}[language=Python, label=code_format, caption={An instance with the unified data format. The triggers that recorded in an item of ``events'' have a coreference relation with each other.}]
{ # one instance
    "id": "instance.001.01",
    "text": "U.S. and British troops were moving on the strategic southern port city of Basra Saturday on Sunday after a massive aerial assault pounded Baghdad at dawn .",
    "events": [
        {
            "type": "attack",
            "triggers": { # triggers that have a coreference relation with each other
                "id": "trigger1",
                "trigger_word": "assault",
                "offset": [22, 23],
                "arguments": [
                    {"mention": "U.S.", "offset": [0, 1], "role": "attacker"},
                    {"mention": "British", "offset": [1 ,2], "role": "attacker"},
                    {"mention": "dawn", "offset": [26, 27], "role": "time"}
                ]
            }
        },
        {
            "type": "motion",
            "triggers": {
                "id": "trigger2",
                "trigger_word": "moving",
                "offset": [5, 6],
                "arguments": [
                    {"mention": "Sunday", "offset": [17, 18], "role": "time"},
                ]
            }
        },
        # .....
    ],
    "event-relations": {
        "temporal": [
            ["trigger1", "before", "trigger2"]
        ],
        "causal": [],
        "subevent": []
    }
}

\end{lstlisting}
\end{minipage}
\end{figure}

\end{document}